# IMAGE DENOISING USING NEW ADAPTIVE BASED MEDIAN FILTER

Suman Shrestha[1, 2]

[1]University of Massachusetts Medical School, Worcester, MA 01655
[2]Department of Electrical and Computer Engineering,
The University of Akron, Akron, OH 44325

## ABSTRACT

*Noise is a major issue while transferring images through all kinds of electronic communication. One of the most common noise in electronic communication is an impulse noise which is caused by unstable voltage. In this paper, the comparison of known image denoising techniques is discussed and a new technique using the decision based approach has been used for the removal of impulse noise. All these methods can primarily preserve image details while suppressing impulsive noise. The principle of these techniques is at first introduced and then analysed with various simulation results using MATLAB. Most of the previously known techniques are applicable for the denoising of images corrupted with less noise density. Here a new decision based technique has been presented which shows better performances than those already being used. The comparisons are made based on visual appreciation and further quantitatively by Mean Square error (MSE) and Peak Signal to Noise Ratio (PSNR) of different filtered images..*

## KEYWORDS

*Impulse Noise, Nonlinear filter, Adaptive Filters, Decision Based Filters.*

## 1. INTRODUCTION

The goal of this study is to examine efficient and reliable image denoising algorithms of impulsive noise. Two algorithms, adaptive filtering techniques and decision based filtering techniques, have been explained in detail while other algorithms have been used for simulation purposes and explained in the review of median filters in Section 2. Further, a technique known as adaptive decision based filtering is proposed and applied which is a new technique for the removal of impulse noise. It is then compared with the previous techniques with respect to their performance.

Images are always preferred to texts in multimedia transmission but all these communications face a common problem: "Noise". One of the most common form of noise is the impulse noise, also known as salt and pepper noise which is caused by unstable voltage which is due to transmission or errors generated in the communication channel. The impulse noise produces fixed values in the pixels which are 0 (pepper noise) and 255 (salt noise). The noise model for an impulse noise can be expressed as:

$$x_i = \begin{cases} 0, & \text{with probability } p_n \\ 255, & \text{with probability } p_p \\ \phi_i, & \text{with probability } 1-(p_n + p_p) \end{cases} \qquad (1)$$

where $x_i$ denotes the pixel of corrupted image, $\phi_i$ denote the pixel, $p_n$ and $p_p$ are the probability of the pixel corrupted with pepper noise and salt noise respectively, where $p_n$ and $p_p$ = 1/2*Noise Ratio, where Noise ratio lies between 0 and 1.

Noise filtering techniques can either be linear or non-linear. The linear filtering technique applies the algorithm linearly to all the pixels in the image without defining the image as corrupted or uncorrupted pixel. Since the algorithm applies to all the pixels in the image, so this causes the uncorrupted pixels to be filtered and hence these filtering techniques are not effective in removing impulsive noises. On the other hand, non-linear filtering technique [1]-[2] is a two phase filtering process. In the first phase, the pixels are identified as corrupted or uncorrupted pixel and in the second phase, the corrupted pixel is filtered using the specified algorithm while the uncorrupted pixel value is retained. The most widely used non-linear filter is the median filter which uses the median value to replace the corrupted pixel, and these filters have the capability to remove impulsive noise while preserving the edges. A diagram highlighting the principle of non-linear filters is shown in Figure 1.

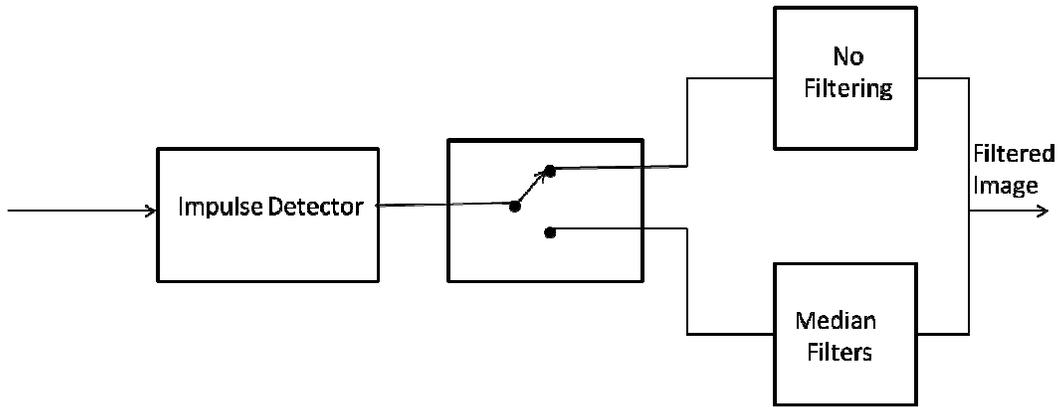

Figure 1. Non-linear Filtering Technique.

Many researchers have been working on different non-linear filtering techniques. In Section 2, some of these filtering techniques will be discussed like the standard median filter (SM), center-weighted median filter (CWMF) [3], tri-state median filter (TSMF [4], progressive switching median filter (PSMF) [5] and adaptive progressive switching median filter (APSMF) [6]. In Section 3, two of the currently known best non-linear filtering techniques, the conventional adaptive filtering technique (AMF) and the decision based median filtering technique (DBMF) will be explained. In Section 4, a new decision based median filter known as the adaptive decision based median filter (which is identified as the combination of the adaptive filter and decision based filter) will be proposed. The simulation results will be shown and discussed in Section 5 and the conclusions will be drawn in Section 6. In this paper, the performance evaluation of the simulated results is carried out by calculating the Peak Signal to Noise Ratio of the filtered image. The peak signal to noise ratio of an image is given by:

$$PSNR = 10 * log_{10} \frac{255^2}{MSE} \qquad (2)$$

where, MSE is the mean square error and is given by:

$$MSE = \frac{1}{MN} \sum_{i=1}^{M} \sum_{j=1}^{N} [I(i,j) - \bar{I}(i,j)]^2 \qquad (3)$$

Here, M and N are the horizontal and the vertical dimensions of the image; I is the original uncorrupted image and $\overline{I}$ is the filtered image.

## 2. REVIEW OF MEDIAN FILTERS

The basic median filter is the standard median filter. In this method, a square window of size 2k+1, where k goes from 1 to N, is used to filter the center pixel. The pixels in the window are first sorted and the center pixel is changed to the median value of the sorted sequence. This method is the simplest of the median filtering techniques and because of its simplicity; it has been used for a long time.

The second type of median filtering is the center weighted median filtering technique. This is similar to the standard median filtering technique except that the center pixel in the window to be filtered is assigned a certain weight, i.e. the center pixel is repeated for some number of times which is defined as the weight. Since the center pixel in the window to be filtered is repeated for some number of times, this technique is called the center weighted median filter. Let, $I_{ij}$ be the center pixel of the window W, then the output (filtered) pixel of the center weighted median filter as given by [3] is:

$$\overline{I}_{ij} = median[I_{i-s,j-t}, w^{\cdot\cdot} I_{ij} \mid (s,t) \in W, (s,t) \neq (0,0)] \qquad (4)$$

where, W is the window, w is the weight of the center pixel, $w^{\cdot\cdot} I_{ij}$ denotes that $I_{ij}$ is repeated w times in calculating the median.

The next median filter used to remove impulsive noise is the tri-state median filter. It uses the standard median filter's and the center weighted median filter's outputs to develop a new filtering technique. The algorithm as developed by \cite{4} is given by:

$$\overline{I}_{ij}^{TSM} = \begin{cases} I_{ij}, & T_D \geq d_1 \\ I_{ij}^{CWM}, & d_2 \leq T_D < d_1 \\ I_{ij}^{SM}, & T_D < d_2 \end{cases} \qquad (5)$$

Here $I_{ij}^{CWM}$ and $I_{ij}^{SM}$ are the outputs of the center weighted median filter and the standard median filter, $d_1 = \mid I_{ij} - I_{ij}^{SM} \mid$ and $d_2 = \mid I_{ij} - I_{ij}^{CWM} \mid$ and $T_D$ is the threshold used which is between 0 and 255.

The median filtering technique, which uses a more complex technique to detect the impulse noises and then filter the pixels, is the progressive median filtering technique [5]. This technique uses an impulse detector technique to detect the pixels corrupted with impulse noise and then uses a filtering technique to filter these corrupted pixels. Both these detection and filtering techniques are applied through several iterations to ensure that all the noisy pixels have be detected (in the case when the image is highly corrupted) and the corrupted pixels have been filtered using this method. The two steps in this method are:

i.  Impulse Noise Detection: This step produces two image sequences, one the pixel values of the image after $N_D$ iterations and the other which indicates whether the pixel is corrupted or uncorrupted. The second image sequence is named as a flag sequence.

Initially, all the pixel values of the flag sequence is assigned as 0 which indicates that the pixel value is uncorrupted. Once the detection algorithm is applied, the flag sequence is changed to 1 if the pixel is detected as noise and is unaltered if it is a noise-free pixel. Let, $I_i^{(n-1)}$ be the image sequence after (n-1) iteration and $I_i^{(n)}$ be the sequence after n iterations, $f_i^{(n-1)}$ be the flag sequence after (n-1) iterations and $f_i^{(n)}$ be the flag sequence after n iterations. A window (W) is specified to calculate the median value of the pixel. The median value of the corresponding pixel of the image sequence is given by:

$$med_i^{(n-1)} = median[I_i^{(n-1)} \mid i \in W] \qquad (6)$$

The impulse noise is then detected using $med_i^{(n-1)}$ and $I_i^{(n-1)}$ and is given by:

$$f_i^{(n)} = \begin{cases} f_i^{(n-1)}, & \text{if } \mid I_i^{(n-1)} - med_i^{(n-1)} \mid < T_D \\ 1, & otherwise \end{cases} \qquad (7)$$

Here, $T_D$ is a predefined threshold. Using this, the value of each of the pixel is then modified as (to be used for next iteration):

$$I_i^{(n)} = \begin{cases} m_i^{(n)}, & \text{if } f_i^{(n)} \neq f_i^{(n-1)} \\ I_i^{(n-1)}, & \text{if } f_i^{(n)} = f_i^{(n-1)} \end{cases} \qquad (8)$$

This is repeated for $N_D$ iterations and gives two image sequences: the final image sequence and the flag sequence.

ii.  Filtering Phase: This phase filters the corrupted pixels obtained from the first phase. It uses two image sequences to filter the image, one is the flag sequence from the first phase and the other is the corrupted pixel sequence. Let, $g_i^0 = f_i^{(N_D)}$ be the flag sequence and $I_i^0$ be the corrupted image sequence. A window (W) is specified to calculate the median value of the pixel. For the $N^{th}$ iteration, the median value of the image sequence from the specified window is calculated using only the good pixel values, i.e.,

$$med_i^{(n-1)} = median[I_i^{(n-1)} \mid g_i^{(n-1)} = 0, i \in W] \qquad (9)$$

After calculating the median value, the image sequence and the flag sequence are modified as:

$$I_i^{(n)} = \begin{cases} m_i^{(n)}, & \text{if } g_i^{(n-1)} = 1; M > 0 \\ I_i^{(n-1)}, & \text{otherwise} \end{cases} \qquad (10)$$

where, M is the number of uncorrupted pixels in the window. If the corrupted pixel is modified, the flag sequence is modified to 0. Thus,

$$g_i^{(n)} = \begin{cases} g_i^{(n-1)}, & \text{if } I_i^{(n)} = I_i^{(n-1)} \\ 0, & \text{if } I_i^{(n)} = med_i^{(n-1)} \end{cases} \qquad (11)$$

This process is repeated until all the pixels of the flag sequence have been changed to 0. Finally, it gives the filtered image $I_{ij}^{N}$, where N is the number of repetition of the filtering technique.

A modification of the progressive switching filtering technique is the adaptive progressive switching filtering technique [6]. This method also filters the image in two phases similar to the previous one. The impulse noise detection phase is similar to progressive switching technique except for Equation 7 in detecting the noise which is applied as:

$$f_i^{(n)} = \begin{cases} f_i^{(n-1)}, & \text{if } \mid I_i^{(n-1)} - med_i^{(n-1)} \mid < T_D; \\ & min(I_i^{(n-1)}) < median_i^{(n-1)} < max(I_i^{(n-1)}) \\ 1, & otherwise \end{cases} \qquad (12)$$

The noise filtering phase is also similar to the previous method except it uses adaptive method to filter the noisy pixels. Adaptive implies that the window size is increased if a certain condition does not meet while filtering the pixel. This means a slight modification is required in Equation 10. The median filtering technique is applied if the number of uncorrupted pixels in the window is greater or equal than half the number of pixels in the window and the window size is less than the maximum specified window size (which is to ensure the correct value of the pixel is obtained) else the window size is increaed by 2 in each of the horizontal and vertical side and the process is repeated, i.e. Equation 10 is modified to:

$$I_i^{(n)} = \begin{cases} m_i^{(n-1)}, & \text{if } g_i^{(n-1)} = 1; \\ & \text{M} > 0.5 * (W * W); \\ & \text{W} < W_{max} \\ W + 2, & otherwise \end{cases} \qquad (13)$$

Here, M is the number of uncorrupted pixels in the window and $W_{max}$ is the maximum window size (predefined) that can be applied. The other steps are similar to the previous method (Progressive Switching Filtering) described above. This is applied until all the corrupted pixels have been filtered.

## 3. CONVENTIONAL ADAPTIVE MEDIAN FILTERING AND DECISION BASED MEDIAN FILTERING

A. Conventional Adaptive Median Filter [7]: The adaptive median filter also applies the noise detection and filtering algorithms to remove impulsive noise. The size of the window applied to filter the image pixels is adaptive in nature, i.e. the window size is increased if the specified condition does not meet. If the condition is met, the pixel is filtered using the median of the window. Let, $I_{ij}$ be the pixel of the corrupted image, $I_{min}$ be the minimum pixel value and $I_{max}$ be the maximum pixel value in the window, W be the current window size applied, $W_{max}$ be the maximum window size that can be reached and $I_{med}$ be the median of the window assigned. Then, the algorithm of this filtering technique completes in two levels as described:

**Level A:**

a) If $I_{min} < I_{med} < I_{max}$, then the median value is not an impulse, so the algorithm goes to Level B to check if the current pixel is an impulse.

b) Else the size of the window is increased and Level A is repeated until the median value is not an impulse so the algorithm goes to Level B; or the maximum window size is reached, in which case the median value is assigned as the filtered image pixel value.

**Level B:**

a) If $I_{min} < I_{ij} < I_{max}$, then the current pixel value is not an impulse, so the filtered image pixel is unchanged.

b) Else the image pixel is either equal to $I_{max}$ or $I_{min}$ (corrupted), then the filtered imaged pixel is assigned the median value from Level A.

These types of median filters are widely used in filtering image that has been denoised with noise density greater than 20%.

B. Decision Based Median Filtering [8]: The decision based median filtering method processes the corrupted image pixels by first detecting whether the pixel value is a corrupted one. This decision is made similar to the one made for adaptive median filtering technique, i.e. based on whether the pixel value to be processed lies between the minimum and the maximum value inside the window to be processed. If the pixel value lies between the minimum and the maximum value in the window, the pixel is left unchanged (as it is detected as a noise free pixel), otherwise the pixel is replaced with the median value of the window or its neighbourhood pixel value. The complete algorithm of the decision based filtering technique is as stated below:

**Step A:** A window is selected and the minimum, maximum and the median value of the window are detected as the first step. Let $I_{ij}$ be the pixel value, $I_{max}$, $I_{min}$ and $I_{med}$ be the maximum pixel value, minumum pixel value and median pixel value respectively in the window.

**Step B:**

a) If $I_{min} < I_{ij} < I_{max}$, then the pixel value is not an impulse noise and hence the same pixel value is retained in the filtered image else the pixel value is an impulse noise.

b) If $I_{ij}$ is an impulse noise, then it is checked to see if the median value is an impulse noise or not. If $I_{min} < I_{med} < I_{max}$ or $0 < I_{med} < 255$, the median is not an impulse noise and the filtered image pixel is replaced by the median value of the window.

c) If the median is also an impulse noise, and in this case the filtered image pixel is replaced by the value of the left neighbourhood pixel value (which has already been filtered).

**Step C:** These steps are repeated until all the pixels have been tested.

## 4. PROPOSED METHOD

A new algorithm (Adaptive Decision Based Median Filtering Algorithm) is implemented to see if the performance of the image is increased with respect to the decision based median filtering. This algorithm seems to be similar to adaptive median filtering technique. A small change is made to this with respect to the decision based median filter and adaptive median filter. The change is made in Step B (c) with respect to decision based median filtering and at the end with respect to adaptive median filtering technique. If the median is also an impulse noise, the window size is increased by 2 in both horizontal and vertical direction (change made with respect to decision based median filter) and the same algorithm is repeated to see if the new median obtained after increasing the window size is an impulse noise. If it is an impulse noise again, the same algorithm is repeated until the maximum size of the window is reached; otherwise the filtered image pixel is replaced by the left neighbourhood pixel value as in the decision based median filtering (change made with respect to adaptive median filter). This new algorithm is applied to see if any better filtering of the corrupted image could be achieved as it is known that adaptive based algorithm provides better filtering than that applied without adaptive median filtering technique.

## 5. SIMULATION RESULTS AND DISCUSSIONS

The above mentioned algorithms are all implemented in two images of size 512 x 512 pixels: the Lena image and the Boat image as shown in the Figures 2 and 3. Impulse noises are added to the images and their performances are evaluated. The noise densities from 5% to 50% are added to the images and their MSE and PSNR are calculated. The PSNR and MSE are calculated from Equations 2 and 3 respectively. It is to be noted that greater the value of PSNR and lower the value of MSE, the filtering technique is better. These results are presented in a tabular form in Table 1 and 2.

Table 1.  PSNR values for Lena Image at different noise densities

| Noise Density | SM | CWMF | TSMF | PSMF | APSMF | AMF | DBMF | ADBMF |
|---|---|---|---|---|---|---|---|---|
| 5% | 34.35 | 35.90 | 39.27 | 34.67 | 41.70 | 38.76 | 45.01 | 44.85 |
| 10% | 33.28 | 33.12 | 35.00 | 31.30 | 39.13 | 37.64 | 41.39 | 41.26 |
| 20% | 28.86 | 25.39 | 25.68 | 27.50 | 35.74 | 35.22 | 37.35 | 36.98 |
| 30% | 23.55 | 19.92 | 19.91 | 24.75 | 33.29 | 32.80 | 34.60 | 34.20 |
| 40% | 19.03 | 16.10 | 15.93 | 22.35 | 31.31 | 30.71 | 32.40 | 32.29 |
| 50% | 15.23 | 13.10 | 12.84 | 20.09 | 29.50 | 28.86 | 30.33 | 30.46 |

Table 2.  PSNR values for Boat Image at different noise densities

| Noise Density | SM | CWMF | TSMF | PSMF | APSMF | AMF | DBMF | ADBMF |
|---|---|---|---|---|---|---|---|---|
| 5% | 30.33 | 32.22 | 34.41 | 34.53 | 33.22 | 34.60 | 40.11 | 40.00 |
| 10% | 29.61 | 30.48 | 32.12 | 32.10 | 32.15 | 33.93 | 37.29 | 36.98 |
| 20% | 26.97 | 24.86 | 25.32 | 28.46 | 30.01 | 32.02 | 33.74 | 33.35 |
| 30% | 22.77 | 19.78 | 19.81 | 26.09 | 28.46 | 29.96 | 31.32 | 30.95 |
| 40% | 18.95 | 15.95 | 15.81 | 23.48 | 27.10 | 28.03 | 29.14 | 28.89 |
| 50% | 15.20 | 13.11 | 12.85 | 20.92 | 25.81 | 26.30 | 27.16 | 27.16 |

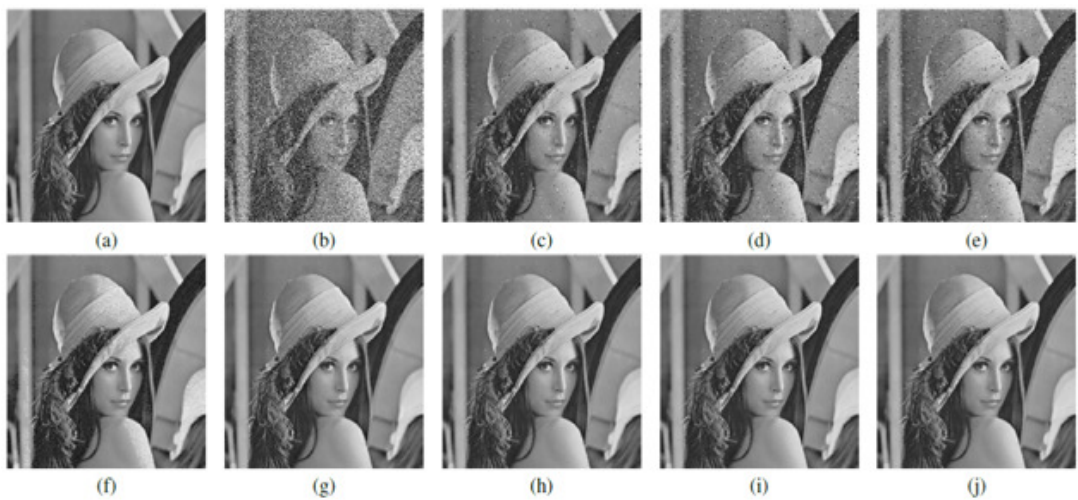

Figure 2. Simulation Results for Lena Image at 50% noise density (a) Original Image, (b) Noisy Image, (c) Standard Median Filtering, (d) Center Weighted Median Filtering, (e) Tri-State Median Filtering, (f) Progressive Switching Median Filtering, (g) Adaptive Progressive Switching Median Filtering, (h) Adaptive Median Filtering, (i) Decision Based Median Filtering, (j) Adaptive Decision Based Median Filtering

Also for the three algorithms described in Section 3 and the proposed method in Section 4, the noise densities are increased from 60% to 90% for the two images and the performances are evaluated. The results of PSNR for these two images are presented in Table 3. The results presented in the tables suggest that adaptive median filtering and decision based median filtering perform better than other techniques. The filtering techniques like standard median filtering, center weighted median filtering and tri-state median filtering (all performed for 3 x 3 window) show good performances only for low noise density. The performances for these filters could be increased if the window size is increased, but as the window size increases, the image tends to show blurring effect. The other filtering techniques like progressive switching median filtering and adaptive progressive switching median filtering show considerably better performance than the above mentioned three filtering techniques at noise density up to 50%. But as the noise density increases further, these techniques lead to lower performance. These two filtering techniques also show considerably low performance if there are salt noise on a white background and pepper noise on a black background in the original image.

Table 3. PSNR values for Boat and Lena Image at higher noise densities for Adaptive, Decision Based and Adaptive Decision Based Filter

| Noise Density | Boat Image | | | Lena Image | | |
|---|---|---|---|---|---|---|
| | AMF | DBMF | ADBMF | AMF | DBMF | ADBMF |
| 60% | 24.81 | 25.51 | 25.78 | 26.57 | 27.72 | 28.36 |
| 70% | 22.28 | 23.43 | 24.16 | 23.67 | 25.61 | 26.57 |
| 80% | 17.97 | 21.26 | 22.42 | 18.46 | 23.12 | 24.58 |
| 90% | 11.70 | 18.14 | 20.61 | 11.93 | 19.61 | 22.25 |

These effects can be seen in Figures 2 and 3 that PSMF fails to remove salt noise from the white background (Figure 2(f)) and pepper noise from the black background (Figure 3(f)). This effect is low (but still exists somewhat) in APSMF (Figures 2(g) and 3(g)).

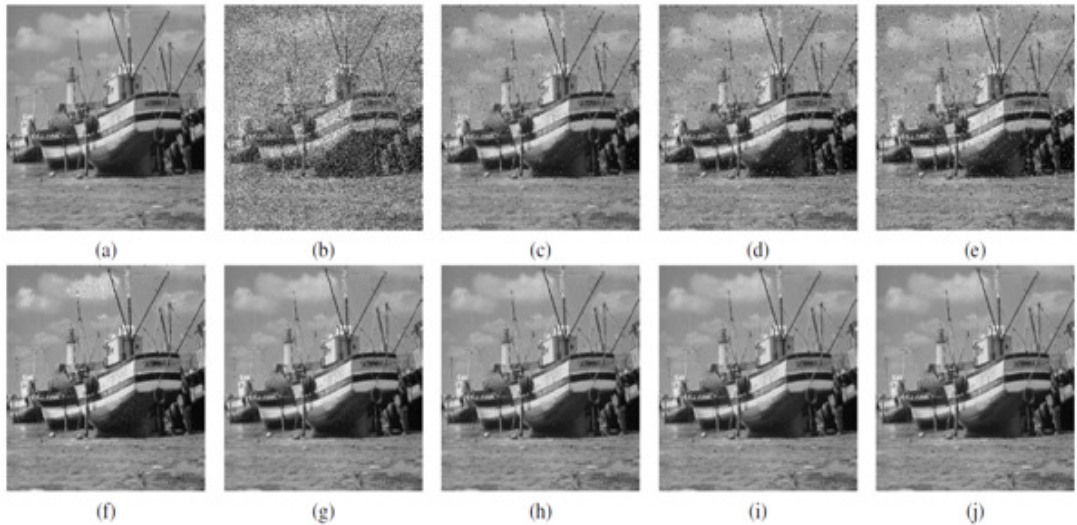

Figure 3. Simulation Results for Boat Image at 30% noise density (a) Original Image, (b) Noisy Image, (c) Standard Median Filtering, (d) Center Weighted Median Filtering, (e) Tri-State Median Filtering, (f) Progressive Switching Median Filtering, (g) Adaptive Progressive Switching Median Filtering, (h) Adaptive Median Filtering, (i) Decision Based Median Filtering, (j) Adaptive Decision Based Median Filtering

When compared to these five filtering techniques, conventional adaptive median filtering and decision based median filtering show better performances overall. The performance also somewhat depends on the image that is being analyzed. In the Lena image, adaptive progressive median filtering shows better performance than the conventional adaptive median filtering technique for noise density up to 50% based on PSNR calculation, but based on the visual effect, the adaptive median filtering has better results. This is because of the salt noise which is yet not filtered on adaptive progressive median filtering technique on the white background. But if the Boat image is used, which has more dark spots, the adaptive progressive switching technique shows degradation in the performance than the adaptive median filtering technique. Also the main disadvantages of the adaptive progressive switching as compared to the traditional adaptive median technique is that it takes relatively high computation time and is not very well suited for real time applications.

On the other hand, decision based median filtering shows the best performance for all noise densities. Upon analysis, it is found that this technique can filter the impulsive noise for densities up to 60% without any blurring effect on the filtered image. But for larger noise densities, although it produces the best results out of the mentioned techniques, the blurring of the filtered images tends to increase. This effect is however a little less if the proposed technique called adaptive decision based median filtering is used. This technique shows performance similar to decision based median filtering with noise densities up to 50%, but as the noise density increases further, the adaptive technique performs better than the decision based technique based on both visual clarity and PSNR calculation.

Thus, based on the analysis presented here, it is clear that adaptive filtering technique shows better performance than the filtering done without any adaptive technique. This can be seen with the comparison of Progressive switching technique to the same technique used with adaptive form and also of the decision based technique to its adaptive form.

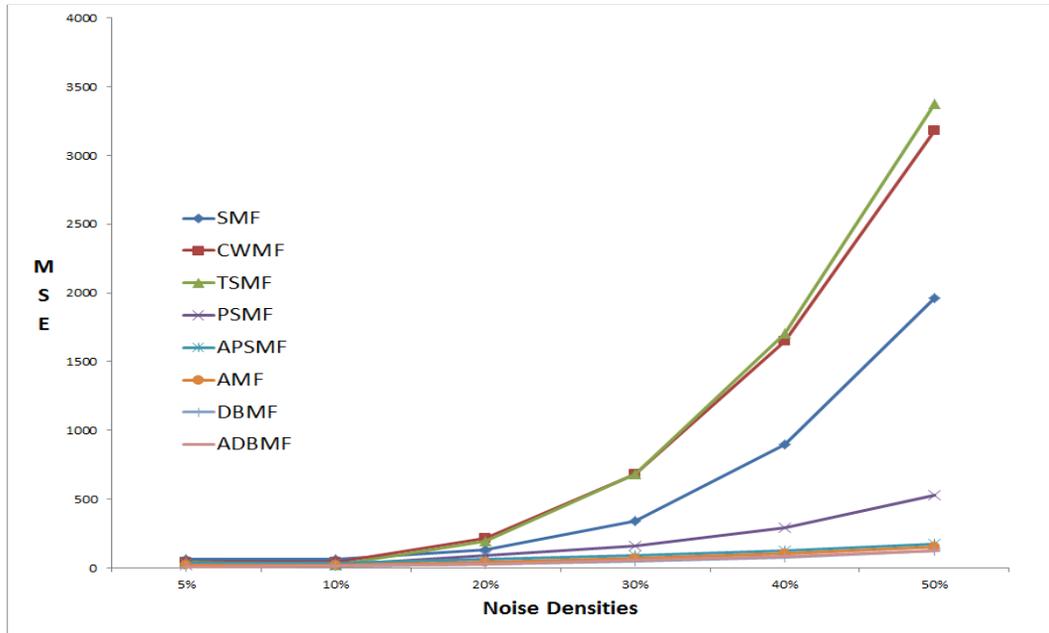

(a)

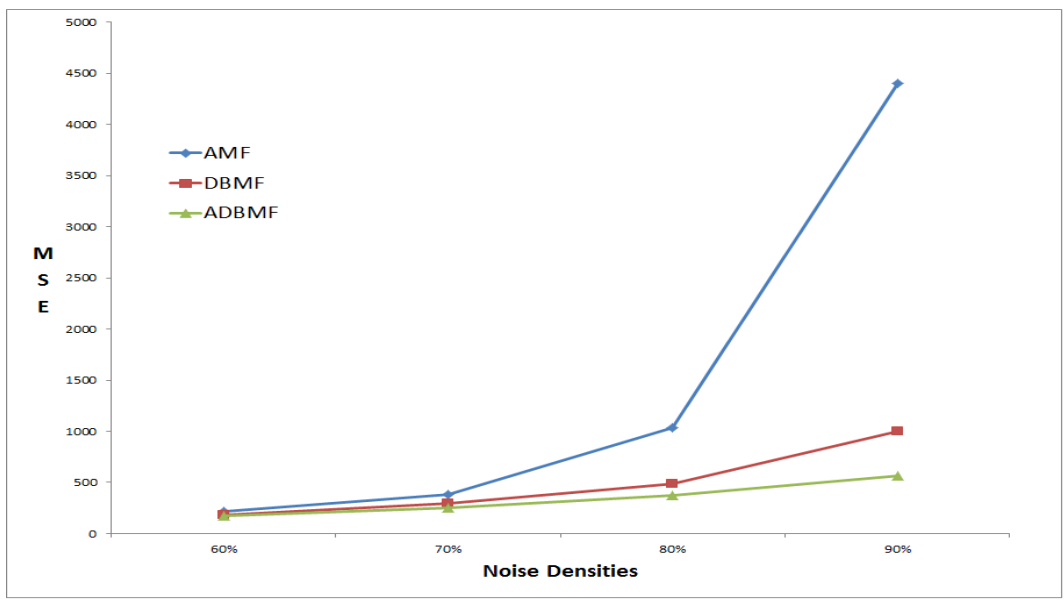

(b)

Figure 4. Plot of MSE for different median filtering techniques for Boat image
(a) MSE Plot for all median filtering techniques
(b) MSE plot for three techniques discussed in Section 3.

When all these filtering techniques are compared, it can be concluded that all these use fixed or variable window size for removal of impulse noise. There is still a need for an algorithm which could automatically decide the window size based on the image and noise density level. Further, the plot of the calculated PSNR and MSE values of the discussed techniques are presented in Figures 4 and 5 (only Boat image) to clarify the discussions.

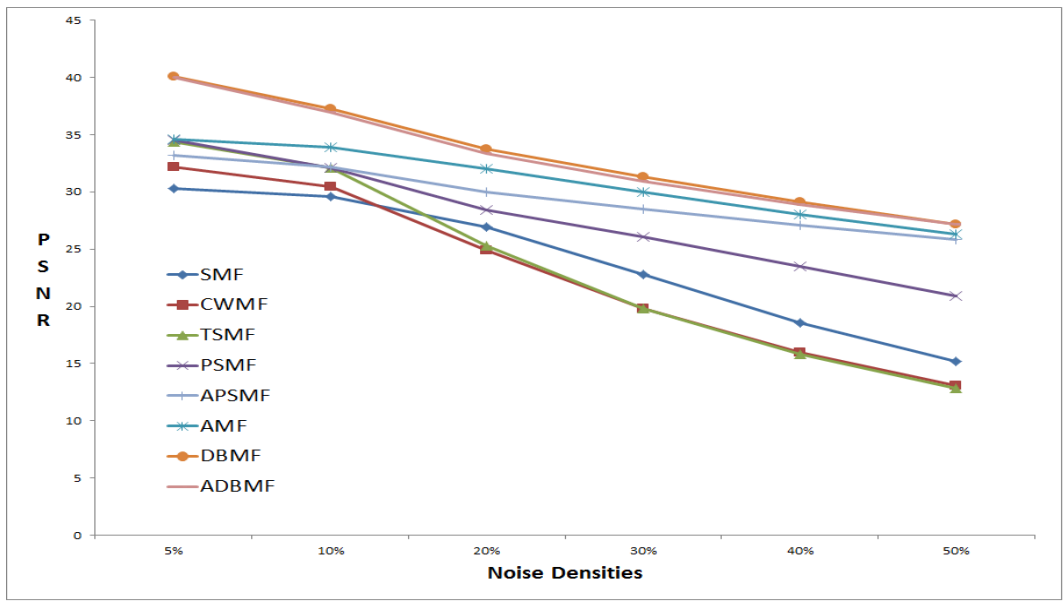

(a)

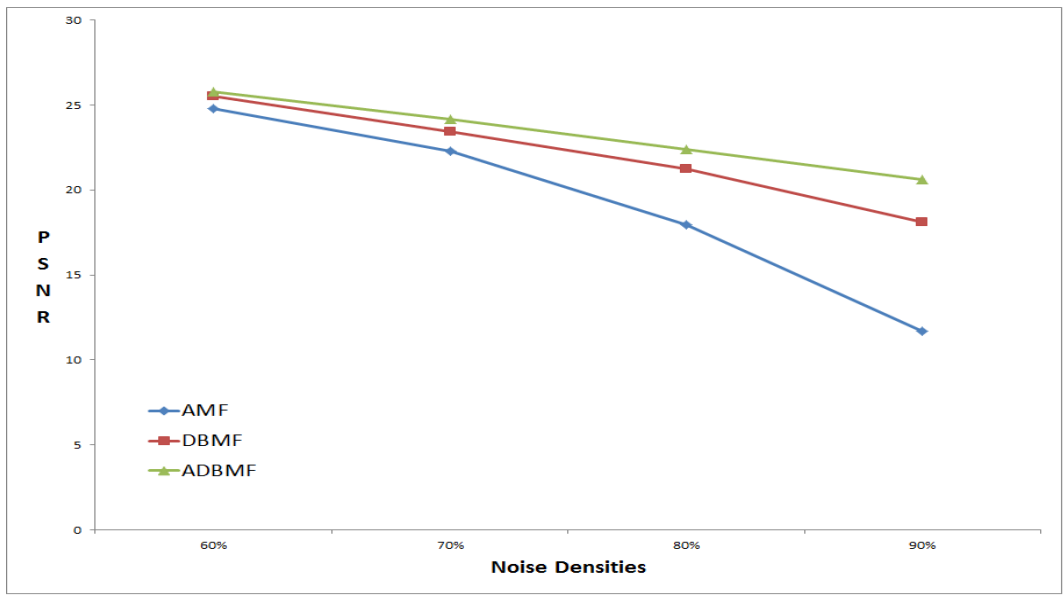

(b)

Figure 4. Plot of PSNR for different median filtering techniques for Boat image
(a) PSNR Plot for all median filtering techniques
(b) PSNR plot for three techniques discussed in Section 3.

In addition to these adaptive median filters, there is also a robust estimation based filter for the removal of high noise density as proposed in [11]. The simulation results provided in the paper can also be applied in hardware and the real implementation of the algorithms could be developed as presented in different literatures [9]-[10].

## 6. CONCLUSIONS

Different median filtering techniques have been discussed and a new technique is proposed which produces the best results among all the techniques available. The results presented above reveal that decision based median filtering technique (among the previous techniques) gives better performance for image denoising of impulsive noise based on PSNR and visual clarity. For very high noise density, adaptive decision based median filtering could be used to get better visual clarity and PSNR values than the decision based median filtering technique. Conventional adaptive median filtering too gives good visual clarity while denoising impulsive noise for noise densities up to 50%. However, in comparison to these techniques, progressive switching and adaptive progressive switching exhibit good performance for low and medium noise densities but takes more computation time and hence is not applicable for real time applications. Thus, as a whole, decision based median filtering technique and adaptive decision based median filtering technique are better techniques than any other techniques described above for filtering impulse noises and the computation time for these techniques are considerably less making them the ideal technique for use in real time application. However for higher noise density, the proposed technique exceeds the decision based technique based on performance. But if the noise densities are considerably higher, then new algorithms must be developed to get much better results than all the techniques described in this paper.

## ACKNOWLEDGEMENTS

The author would like to thank Prof. Dr. S.I. Hariharan (The University of Akron) for his valuable input in this work. This work was performed while the author was pursuing his graduate studies at the University of Akron.

**AUTHOR**

**Suman Shrestha** received his B. E. in Electronics and Communication Engineering in 2009 from Tribhuvan University, Kathmandu, Nepal and his M.S. degree in Electrical Engineering from the University of Akron.

He is currently working as a Research Engineer in the Radiologic Physics Laboratory at the University of Massachusetts Medical School. His work is focused in the design of new X-ray detectors for use in Medical Imaging and Radiotherapy. His research interests are in the field of Polarimetry, Image Processing, Medical Imaging and Radiation and X-ray detectors.